\documentclass[sigconf]{acmart}
\usepackage{multirow}
\usepackage{threeparttable}
\usepackage[normalem]{ulem} 
\usepackage{bbm}
\usepackage{enumitem}
\usepackage{algorithm}
\usepackage{amsmath,algpseudocode}
\usepackage{graphicx}
\usepackage{booktabs}
\usepackage[font=small]{caption}

\usepackage{amssymb}  
\usepackage{amsthm}
\newtheorem{assumption}{Assumption}

\usepackage{graphicx}
\usepackage[caption=false,font=footnotesize]{subfig}

\algnewcommand{\LineComment}[1]{\State \(\triangleright\) #1}

\usepackage[bookmarksnumbered,unicode]{hyperref}
\hypersetup{
    colorlinks=true,
    linkcolor=blue,
    citecolor=red,
    urlcolor=magenta
}

\AtBeginDocument{%
  \providecommand\BibTeX{{%
    \normalfont B\kern-0.5em{\scshape i\kern-0.25em b}\kern-0.8em\TeX}}}

\setcopyright{acmlicensed}
\copyrightyear{2025}
\acmYear{2025}
\acmDOI{}

\acmConference[]{KDD 2026}{August}{2026}
\acmISBN{}

\begin{document}

\title{Generative Large-Scale Pre-trained Models for Automated Ad Bidding Optimization}



\author{Yu Lei}
\affiliation{%
  \institution{Beijing University of Posts and Telecommunications}
  \city{Beijing}
  \country{China}
}
\email{leiyu0210@gmail.com}

\author{Jiayang Zhao}
\affiliation{%
  \institution{Meituan}
  \city{Beijing}
  \country{China}
}
\email{zhaojiayang02@meituan.com}

\author{Yilei Zhao}
\authornote{Corresponding author.}
\affiliation{%
  \institution{Meituan}
  \city{Beijing}
  \country{China}
}
\email{zhaoyilei@meituan.com}

\author{Zhaoqi Zhang}
\affiliation{%
  \institution{Nanyang Technological University}
  \country{Singapore}
}
\email{zhaoqi001@e.ntu.edu.sg}

\author{Linyou Cai}
\affiliation{%
  \institution{Meituan}
  \city{Beijing}
  \country{China}
}
\email{cailinyou@meituan.com}

\author{Qianlong Xie}
\affiliation{%
  \institution{Meituan}
  \city{Beijing}
  \country{China}
}
\email{xieqianlong@meituan.com}

\author{Xingxing Wang}
\affiliation{%
  \institution{Meituan}
  \city{Beijing}
  \country{China}
}
\email{wangxingxing04@meituan.com}

\renewcommand{\shortauthors}{Yu Lei, et al.}
\newcommand{\zhaoqi}[1]{\textcolor{red}{#1}}

\begin{abstract}
Modern auto-bidding systems are required to balance overall performance with diverse advertiser goals and real-world constraints, reflecting the dynamic and evolving needs of the industry. Recent advances in conditional generative models, such as transformers and diffusers, have enabled direct trajectory generation tailored to advertiser preferences, offering a promising alternative to traditional Markov Decision Process-based methods. However, these generative methods face significant challenges, such as the distribution shift between offline and online environments, limited exploration of the action space, and the necessity to meet constraints like marginal Cost-per-Mille (CPM) and Return on Investment (ROI). To tackle these challenges, we propose \textsc{GRAD} (Generative Reward-driven Ad-bidding with Mixture-of-Experts), a scalable foundation model for auto-bidding that combines an Action-Mixture-of-Experts module for diverse bidding action exploration with the Value Estimator of Causal Transformer for constraint-aware optimization. Extensive offline and online experiments demonstrate that GRAD significantly enhances platform revenue, highlighting its effectiveness in addressing the evolving and diverse requirements of modern advertisers. Furthermore, GRAD has been implemented in multiple marketing scenarios at Meituan, one of the world’s largest online food delivery platforms, leading to a 2.18\% increase in Gross Merchandise Value (GMV) and 10.68\%
increase in ROI.

\end{abstract}







\maketitle

\section{Introduction}

Online advertising has become a vital revenue engine for internet platforms, spurring the development of intelligent ad delivery systems. As the online advertising market continues to expand, advertisers are raising their expectations for both the effectiveness and efficiency of their campaigns~\cite{evans2009online}. A core competency of advertising platforms lies in how intelligently their auto-bidding systems help advertisers maximize marketing value under constraints such as budgets and performance objectives~\cite{yang2019aiads,wen2022cooperative,wang2023adversarial,liu2024bird}. In practical industrial scenarios, advertisers exhibit highly heterogeneous and dynamic preferences~\cite{barford2014adscape,liu2025sigma} -- some prioritize brand exposure while others focus on conversion rates or cost efficiency~\cite{lee2012estimating}. Consequently,  auto-bidding systems must optimize general performance~\cite{aggarwal2024auto, deng2021towards} while accommodating flexible support for a wide range of objectives and constraints~\cite{ou2023survey}, adapting swiftly to shifting business demands and market environments.

Recent advances in reinforcement learning~\cite{jin2018real,ye2019deep,mou2022sustainable} and generative modeling~\cite{li2024gas,liu2025deepfake} have significantly transformed the landscape of bidding algorithms, which empower the development of more intelligent~\cite{bijalwan2025unveiling}, adaptive, and personalized solutions tailored to complex and varied industrial needs of modern advertisers~\cite{jiang2025automated}. In particular, conditional generative approaches, leveraging robust architectures such as transformers~\cite{li2024gas,jiang2025optimal,lei2024generative,zhang2025diffusion} and diffusion models~\cite{guo2024generative,zhang2506timemaster}, enable direct generation of bidding actions or entire bidding trajectories by encoding advertiser preferences as vectorized conditions.  This paradigm bypasses the limitations of traditional Markov Decision Process (MDP) frameworks and allows for more flexible real-time decision-making. Inspired by the success of large-scale foundation models like GPT-4~\cite{achiam2023gpt}, DeepSeek~\cite{liu2024deepseek}, and Stable Diffusion~\cite{rombach2022high}, the auto-bidding field is now moving towards foundation model-based methods, where universal decision policies are learned from massive datasets, enabling scalable and adaptive bidding systems capable of addressing a broad spectrum of advertiser needs.

Despite their promise, generative auto-bidding systems face two fundamental challenges rooted in real-world industrial deployment: \textbf{(1) Offline–online distribution mismatch.} There exists a significant gap between offline training—based on static historical data—and the dynamic nature of online environments. This mismatch often leads to performance degradation under distribution shifts (e.g., sudden changes in user behaviour or competitive intensity) and long-tail scenarios (e.g., holiday traffic spikes or cold-start campaigns). In such cases, the bidding system may converge to suboptimal policies or become unstable due to limited exposure to unseen states and insufficient exploration. \textbf{(2) Action space design under hard constraints.} Defining an adequate bidding action space is inherently challenging. While outperforming historical strategies requires exploring novel, unobserved actions, unconstrained exploration may violate strict business constraints—such as budget pacing or delivery deadlines—leading to revenue loss or resource overspend. This trade-off between strategic innovation and deployment safety necessitates architectural innovations that can balance exploration with constraint adherence.

To address these challenges fundamentally, we propose \textsc{GRAD} (\textbf{G}enerative \textbf{R}eward-driven \textbf{AD}-bidding), a novel and scalable generative auto-bidding architecture. Built upon the principle of the scaling law—where model performance scales exponentially with computational resources—\textsc{GRAD} is designed to achieve both adaptability and deployment reliability in industrial auto-bidding systems. This architecture provides a robust solution to offline-online discrepancy and suboptimal action space limitations through synergistic innovations. Specifically, the Action-MoE Module employs a Mixture-of-Experts (MoE) mechanism~\cite{huang2025moe} for constrained exploration in high-dimensional action spaces, dynamically generating novel strategies beyond historically observed actions while inherently mitigating feature distribution shifts through domain-informed constraint simulation. Complementing this, the Causal Transformer Value Estimator performs counterfactual inference to evaluate the reward of unexecuted actions under complex advertiser constraints, enabling precise and objective-aligned optimization. Together, this decoupled exploration-optimization paradigm—powered by scalable MoE and deep causal reasoning—has been successfully deployed across Meituan's advertising platform, demonstrating a sustainable pathway for reliable generative auto-bidding in real-world industrial environments.

The main contributions of the paper are summarized as follows:
\begin{enumerate}[leftmargin=*]
    \item We propose a novel end-to-end auto-bidding framework, GRAD, which supports expert configuration and deployment according to online environment resources, providing a more practical and flexible deployment solution for real-world advertising systems.
    \item We introduce the ActionMoE module to enhance exploration efficiency by utilizing data from randomly perturbed bids (increases or decreases), enabling the model better to uncover optimal bidding strategies in a high-dimensional action space.
    \item We conduct large-scale comparative experiments and online A/B tests on Meituan's advertising platform, GRAD achieved statistically significant improvements: +2.18\% GMV and +10.68\% ROI over baseline methods.
\end{enumerate}

\section{Preliminary}
\subsection{Auto-Bidding Problem}\label{sec:problem}
An auto-bidding environment is commonly formalized as a sequence of $I$ discrete auction rounds. In each round, the advertiser strategically selects a bid within a repeat auction framework. An impression is assigned to the advertiser if their bid exceeds that of all competitors, incurring a cost determined by the underlying auction mechanism. The primary objective is to maximize the aggregate valuation of all impressions acquired across the auction horizon:
\begin{equation}
\text{maximize} \quad \sum\nolimits_{i=1}^{I} x_i v_i
\end{equation}
Here, the $i$ represents of the $i$-th round, $x_i \in \{0,1\}$ indicates impression outcome (1 if won, 0 otherwise), and $v_i \in \mathbb{R}^+$ represents the impression value, typically derived via Click-Through Rate (CTR) or Conversion Rate (CVR).

In addition to value maximization, auto-bidding systems must satisfy various constraints to ensure campaign feasibility. Among these, the budget constraint is a primary hard constraint that is enforced in real time. In contrast, performance-based key performance indicators (KPIs), such as Cost Per Click (CPC), depend on post-auction user feedback, rendering them inherently non-causal and more appropriately modeled as soft constraints. In this work, we focus on the CPC constraint as a representative example and formalize the overall bidding objective as follows:
\begin{equation}\label{equ:problem1}
\begin{aligned}
\text{maximize} \quad & \sum\nolimits_{i=1}^{I} x_i v_i \\
\text { s.t. }  \quad & \sum\nolimits_{i=1}^{I} x_i c_i \leq B \\
                \quad & \frac{\sum_{i=1}^{I} x_i c_i}{\sum_{i=1}^{I} x_i p_i} \leq C
\end{aligned}
\end{equation}
Here, $B$ and $C$ correspond to the total budget and the upper bound on CPC, respectively, while $c_i$ denotes the cost and $p_i$ denotes the predicted CTR associated with the $i$-th impression.

The optimization problem above involves a high-dimensional decision space and uncertain auction outcomes. Previous work~\cite{he2021unified,su2024auctionnet} reformulates it as a linear program and derives closed-form bidding rules. In the presence of a single CPC constraint, the optimal bidding $b^*_i$ can be calculated by:
\begin{equation}\label{equ:bi}
b_{i}^* = \lambda_0^* v_{i} + C \lambda_1^* p_{i}
\end{equation}
Here, $\lambda^* = \lambda_0^* + C \lambda_1^*$ acts as a unified control variable for adaptive bidding. In practice, CPC constraint violations are mitigated by adding penalty terms to the training objective, as detailed in Section~\ref{sec:opt}.

\subsection{Auto-bidding Decision Process}
\label{sec:opt}

Online advertising environments are highly dynamic and characterized by significant temporal volatility in user behavior, auction competition, and budget consumption. These challenges require continuous and adaptive optimization of bid parameters to ensure sustained campaign performance and long-term value maximization. As a result, auto-bidding must be approached as a sequential decision-making problem, where each decision can have far-reaching effects on future outcomes.

To address this complexity, we adopt the Transformer paradigm~\cite{chen2021decision,roberts2024powerful,hu2023graph} to model sequential dependencies in the bidding process, which we formalize as a MDP. The MDP is specified by the tuple $(\mathcal{S}, \mathcal{A}, \mathcal{T}, \mathcal{R}, \gamma)$ 
where $\mathcal{S}$, $\mathcal{A}$, $\mathcal{T}$, and $\mathcal{R}$ denote the state space, action space, state transition function, and reward function, respectively, and $\gamma$ is the discount factor. At each decision cycle $t$, the agent observes the current state $s_t \in \mathcal{S}$, selects an action $a_t \in \mathcal{A}$, transitions to the next state $s_{t+1} = \mathcal{T}(s_t, a_t)$ according to the environment dynamics, and receives an immediate reward $r_t$ that quantifies the value generated by the action. 

As described in Eq. ~\eqref{eq-mdp}, the policy $\pi: \mathcal{S} \to \Delta(\mathcal{A})$ is optimized to maximize the expected cumulative reward over the campaign horizon, enabling the agent to learn effective bidding strategies from sequential interactions with the auction environment.

\begin{equation}
\label{eq-mdp}
\pi^* = \arg\max_{\pi}~\mathbb{E}_{\pi} \left[ \sum_{t=0}^{T} \gamma^t r_t \right]
\end{equation}

The main components of the MDP are specified as follows:

\begin{itemize}[leftmargin=*]
    \item \textbf{State} ($s_t$): A multidimensional vector that comprehensively characterizes the campaign’s current status. Key features include the temporal horizon (e.g., elapsed and remaining time in the campaign), residual budget, expenditure velocity (rate of budget consumption), and the achievement ratio of KPIs relative to the constraint $C$ as defined in Eq.~\eqref{equ:problem1}.
    
    \item \textbf{Action} ($a_t$): A vector representing the differential adjustments to the bidding coefficients for each constraint. These adjustments directly influence the bidding strategy by modulating the weights assigned to value and constraint terms.
    
    \item \textbf{Reward} ($r_t$): The immediate value accrued from impressions obtained between decision cycles $t-1$ and $t$. This reward is typically measured as the sum of the realized value (e.g., clicks, conversions, or revenue) generated by the set of impressions acquired in that interval, providing direct feedback to guide policy optimization.
\end{itemize}

\section{Method}
\begin{figure*}[htbp]
    \centering
    \includegraphics[width=0.9\linewidth]{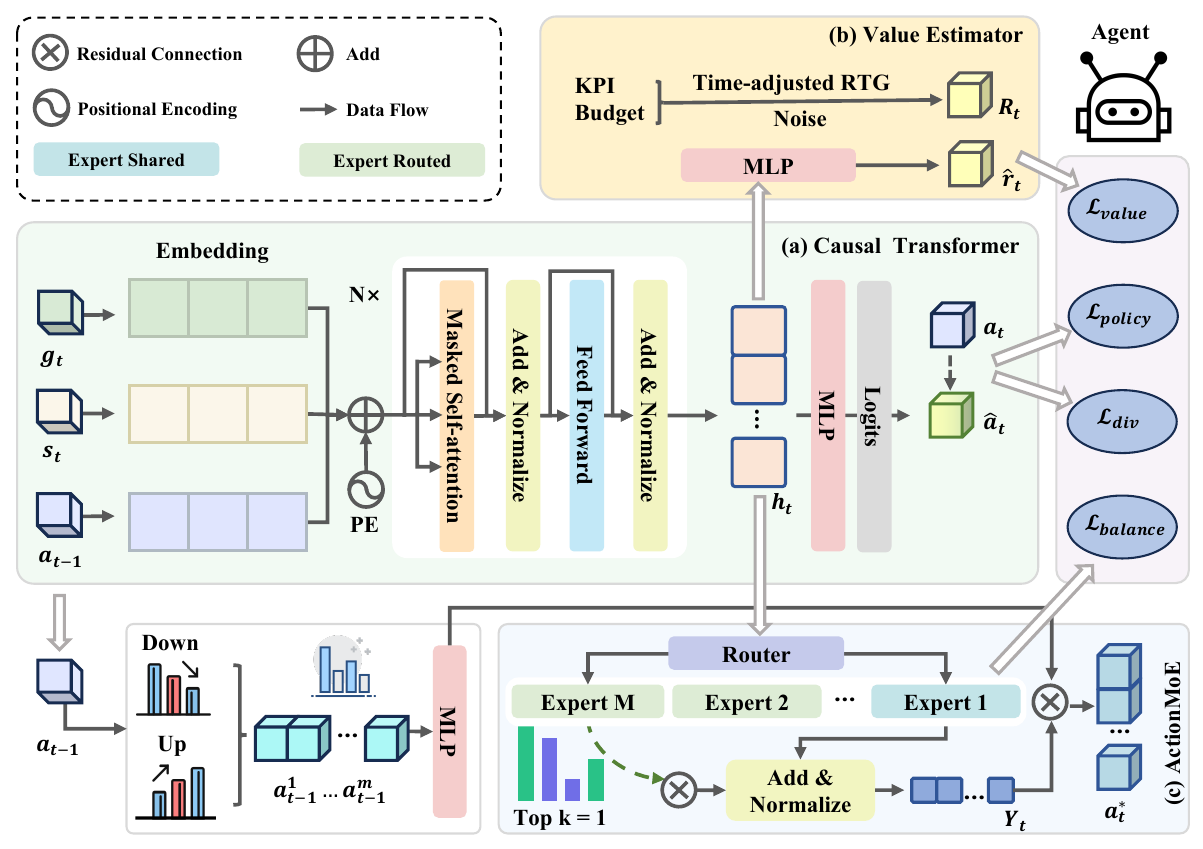}
    \vspace{-2mm}
    \caption{Overall architecture: (a) a Causal Transformer module based on the decoder-only architecture of Transformer(see §\ref{sec:causal_transformer}); (b) a Value Estimator (see §\ref{sec:value_estimator}); (c) an Action MoE module, as described in §\ref{sec:actionmoe}.
}
    \label{fig:method}
\end{figure*}

The proposed framework, GRAD, establishes a novel foundation for adaptive auto-bidding,  which is visually detailed in Figure~\ref{fig:method}. GRAD employs a Causal Transformer (CT)~\cite{chen2021decision} to model complex temporal dependencies in state-action histories. To address critical challenges in online advertising, it integrates two specialized modules: ActionMoE , which enhances exploration efficacy and policy robustness through a learnable MoE mechanism; and the Value Estimator, which incorporates explicit constraints (e.g., budget pacing and CPC targets) directly into the reward signal. This synergistic combination enables the GRAD framework to effectively balance exploration-exploitation trade-offs, dynamically satisfy complex operational constraints, and generate high-quality bidding decisions in real-time ad bidding.


\subsection{Causal Transformer}
\label{sec:causal_transformer}
The CT models temporal dependencies through causal attention mechanisms, enabling generative decision-making for auto-bidding. The action-generation policy is defined as:
\begin{align}
    \hat{a}_t & \sim \pi\big(\cdot \mid \{s_{\leq t}, a_{< t}, g_t\}\big) = \mathrm{CT}\big(\cdot \mid \{s_{\leq t}, a_{< t}, g_t\}\big)
    \label{eq:policy}
\end{align}
where $\hat{a}_t$ is the predicted action at timestep $t$, $s_{\leq t}$ denotes the state history up to $t$, $a_{< t}$ represents the action history prior to $t$, and $g_t$ is the \textit{return-to-go} (RTG) that represents cumulative future rewards:
\begin{align}
    g_t &= \sum_{\tau=t}^{T} r_\tau, \quad r_\tau = x_\tau v_\tau 
    \label{eq:rtg}
\end{align}
where $r_\tau$ is the immediate reward at $\tau \in T$ , $x_\tau \in \{0,1\}$ indicates impression outcome (1 if won, 0 otherwise), and $v_\tau$ is the advertisement value.

To construct the input representation for the CT, we embed the principal features $g_t$, $s_t$, and $a_{t-1}$ and incorporate positional encoding to capture temporal dynamics. The final input vector $\mathbf{h}_t^{(0)}$ at timestep $t$ is formulated as:

\begin{align}
    \mathbf{h}_t^{(0)} &= \text{LayerNorm}\Big( E_g(g_t) \oplus E_s(s_t) \oplus E_a(a_{t-1}) \oplus \text{PE}(t) \Big) 
    \label{eq:embedding}
\end{align}
where $\oplus$ denotes vector concatenation, $E_g$, $E_s$, and $E_a$ are linear embedding layers for the RTG, state, and previous action, respectively, $\text{PE}(t)$~\cite{kazemnejad2023impact} denotes the positional encoding at timestep $t$.


The model employs $N$ stacked transformer blocks to extract high-level representations from the input sequence. Formally, the hidden state at time step $t$ and block $n$ is computed as:
\begin{align}
    \mathbf{h}_t^{(n)} = \text{Block}^{(n)}\big(\mathbf{h}_t^{(n-1)}\big), \quad n = 1,\dots,N,
    \label{eq:blocks}
\end{align}
where $\text{Block}^{(n)}$ denotes the $n$-th transformer block.
On top of the final-layer features $\mathbf{h}_t^{(N)}$, dual prediction heads are employed to generate task-specific outputs. In particular, the policy head produces the predicted action $\hat{a}_t$ as follows:
\begin{align}
    \hat{a}_t = \tanh\big(\text{MLP}_{\text{policy}}(\mathbf{h}_t^{(N)})\big)
\end{align}
Mean squared error (MSE) loss is used to optimize the policy head at each time step:
\begin{align}
    \mathcal{L}_{\text{policy}} = \frac{1}{T} \sum_{t=1}^{T} \big\|\hat{a}_t - a_t\big\|_2^2,
    \label{eq:loss_policy}
\end{align}
where $T$ is the sequence length and $a_t$ denotes the ground-truth action at time step $t$.

\subsection{Value Estimator}
\label{sec:value_estimator}

To mitigate the \textbf{offline-online distribution mismatch} induced by environmental nonstationarity in real-time bidding systems, we propose a Value Estimator that utilizes the precomputed hidden states $\mathbf{h}_t$ from Eq.~\eqref{eq:blocks} to enhance action strategy learning. We introduce the Value Estimator in detail as follows:

\noindent (1) \textbf{Reward from Prediction.}  This component is the value prediction generated from $\mathbf{h}_t$: 
\begin{align}
    \hat{r}_t &= \text{MLP}_{value}(\mathbf{h}_t) 
\label{eq:value_head}
\end{align}

This initial estimate $\hat{r}_t$ serves as the foundation for subsequent dynamic adjustment, reflecting the model's current understanding of the environment and providing basis for policy optimization. 

\noindent (2) \textbf{Reward with Time-Adjustment.} 
Conventional reward estimators in computational advertising often suffer from myopic optimization and time-agnostic modeling, leading to offline-online distribution mismatch. To mitigate end, we design a reward signal $R_t$ that explicitly incorporates temporal and budgetary factors to guide and regularize $\hat{r}_t$:
\begin{equation}
     R_t = \Gamma(t) \cdot \Omega(t) \cdot \Pi(t) \cdot g_t + \Theta_{noise}
    \label{eq:modular_dynamic_value}
\end{equation}

where $g_t$ denotes cumulative future rewards at time $t$ from Eq.~\eqref{eq:rtg}, other modules are defined as follows:
\begin{itemize}[leftmargin=*,noitemsep]
    \item Temporal Decay: 
    $\Gamma(t) = e^{t}$ models the diminishing reward sensitivity over the campaign duration.
    \item Cost Penalty: 
    $\Omega(t)= \min\left(1, \left(\frac{C}{\mathrm{CPC}_t}\right)^\gamma\right)$ downweights rewards when the $\mathrm{CPC}_t$ exceeds the constraint $C$ from Eq. \eqref{equ:problem1}. The $\gamma \in [1, +\infty)$ controls the strength of cost penalty:
    
    \item Budget Efficiency: $\Pi(t)$ is the remaining budget proportion, dynamically reflecting expenditure trajectory. Its evolution follows:
    \begin{equation}
        \frac{\partial \Pi(t)}{\partial t} \propto -\mathrm{CPC}_t
    \end{equation}
    \item Stochastic Noise:  $\Theta_{noise} \sim \mathcal{N}(0,\sigma^2)$ accounts for market volatility. $\mathcal{N}$ denotes the normal distribution.
\end{itemize}

\noindent (3) \textbf{Objective Design.} During training, the predicted value $\hat{r}_t$ is updated under the supervision of $R_t$, enabling the model to capture both immediate and long-term objectives under practical constraints. We minimize temporal-decoupled MSE with increasing weights to counteract late-stage distribution drift:
\begin{align}
\label{equ:value}
    \mathcal{L}_{\text{value}} = \frac{1}{T}
    {\sum_{t=0}^{T} \|\hat{r}_t - R_t\|_2^2}
\end{align}

\subsection{ActionMoE}
\label{sec:actionmoe}

To tackle the challenge of \textbf{action space design under hard constraints}, particularly in complex or high-dimensional environments where conventional exploration may be inadequate or unsafe, we propose ActionMoE—a lightweight exploration module for transformer-based policies. Drawing on DeepSeekMoE~\cite{dai2024deepseekmoe} and Switch Transformers~\cite{fedus2022switch}, ActionMoE employs fine-grained expert segmentation and shared expert isolation. Through a mixture-of-experts and adaptive gating, it enhances exploration, increases action diversity, and maintains stable, safe training~\cite{song2024germ,mu2025comprehensive,dey2024reinforcement}. As depicted in Figure~\ref{fig:method}, the ActionMoE can be organisazed as follows:

\noindent (1) \textbf{Action Exploratory.} 
We generate the scaled action candidates ${\mathbf{a}_{t-1}^m}$ by stochastically perturbing the previous action $\mathbf{a}_{t-1}$ via element-wise scaling to facilitate exploration within the action space through element-wise scaling:
\begin{equation}
\mathbf{a}_{t-1}^m = \mathbf{a}_{t-1} \odot \mathbf{f}_{t-1}^m, \quad \mathbf{f}_{t-1}^m \sim \mathcal{U}[0.8, 1.2)
\end{equation}
Each scaling factor $\mathbf{f}_{t-1}^m$ is sampled independently from a uniform distribution, with identical and independent sampling across all batch elements, sequence positions, and expert indices $m \in \{1, \dots, M\}$, and the operator $\odot$ denotes element-wise multiplication.

\noindent (2) \textbf{Experts Fusion.} We process hidden states $\mathbf{h}_t$ using our Mixture-of-Experts (MoE) module through two complementary pathways: a persistent shared expert that ensures stable learning and preserves baseline performance, and top-1 routed specialized experts that enable efficient and diverse exploration through conditional activation. This integrated design achieves computational efficiency via sparse routing while maintaining linear parameter scaling with transformer depth. Furthermore, by balancing robust exploitation (through the shared expert) and targeted exploration (via specialized experts), we effectively optimize the exploration-exploitation trade-off in complex, high-dimensional environments.

The architecture employs dual processing pathways: a persistent shared expert, defined as follows, to ensure stable learning and baseline performance preservation: 
\begin{equation}
    \mathbf{h}_t^{\text{shared}} = \operatorname{FFN}_{\text{shared}}(\mathbf{h}_t)
    \label{eq:shared}
\end{equation}
which ensures stable learning and baseline performance preservation. 

Complementing this, a specialized pathway activates conditionally selected experts through top-1 routing. Expert selection is determined by
\begin{equation}
\begin{cases}
    \text{gate}_t^{(m)} = \mathbb{I}[m = m^*] \\[1.0em]
    m^* = \arg\max_{m} \left( \operatorname{Softmax}(\mathbf{h}_t^\top \mathbf{e}_m) \right)
\end{cases}
\label{eq:expert_select_gating}
\end{equation}
where \(m^*\) identifies the chosen expert and \(\mathbf{e}_m\) represents trainable routing embeddings. This mechanism, implementing a gating function, activates a single expert per input. The resulting routed output is computed as:
\begin{equation}
    \mathbf{h}_t^{\text{router}} = \sum_{m=1}^{M} \text{gate}_t^{(m)} \cdot \operatorname{FFN}_{\text{routed}}^{(m)}(\mathbf{h}_t)
    \label{eq:routed_output}
\end{equation}

The final exploratory action output combines both pathways through:
\begin{equation}
    \mathbf{Y}_t = \mathrm{LayerNorm} \left( \mathbf{h}_t^{\text{shared}} + \mathbf{h}_t^{\text{router}} \right)
    \label{eq:final_output}
\end{equation}

This design delivers significant advantages: the persistent expert (Eq.~\ref{eq:shared}) stabilizes gradients while the conditional activation (Eq.~\ref{eq:expert_select_gating}) enables targeted exploration. Computational efficiency stems from top-1 routing, and parameter count scales linearly with transformer depth \(N\), maintaining exploitation fidelity while expanding exploratory capabilities for complex decision environments.

After obtaining the processed  output $\mathbf{Y}_t$ subsequently undergoes processing through a multi-layer perceptron, whose outputs form residual connections with the candidate actions $\{\mathbf{a}_{t-1}^m\}_{m=1}^M$ to generate the final weighted actions:
\begin{equation}
    \mathbf{a}_t^{\ast} = \text{MLP}(\mathbf{Y}_t) + \sum_{m=1}^M \omega_m \mathbf{a}_{t-1}^m
\end{equation}
where $\omega_j$ represent trainable weighting coefficients, and $\mathbf{a}_t^m $ are $m$ candidate actions at timestep $t$. This residual formulation enables adaptive combination of transformed state features and candidate action embeddings to produce the final weighted action set $\{ \omega_1\mathbf{a}_t^1, \dots, \omega_m\mathbf{a}_{t-1}^m \}$.

The overall workflow of ActionMoE is summarized in Algorithm~\ref{alg:forward_pass}:

\begin{algorithm}[H]
\caption{ActionMoE Exploration Module Forward}
\label{alg:forward_pass}
\begin{algorithmic}[1]
\Require State representation $\mathbf{h}_t$, nominal action $\hat{a}_t$
\Ensure Refined action ensemble $\mathbf{a}_t^{\ast}$

\State Compute MoE output:
\State $\quad \mathbf{Y}_t \gets \operatorname{ActionMoE}(\mathbf{h}_t)$ \Comment{Sec~\ref{sec:actionmoe}}
\State Generate residual vector:
\State $\quad \mathbf{U}_t \gets \operatorname{MLP}(\mathbf{Y}_t)$ \Comment{$\mathbf{U}_t$}
\State Construct candidate actions:
\State $\quad \{\mathbf{a}^m_{t-1}\}_{m=1}^M \gets f(\hat{a}_t)$ \Comment{Candidate generation function}
\State Apply residual connection:
\For{$m \gets 1$ \textbf{to} $M$}
\State $\quad\mathbf{a}_t^{\ast} \gets \omega_m \mathbf{a}^m_{t-1} + \mathbf{U}_t$ \Comment{$\omega_m $: learnable weights}
\EndFor
\State \Return $\mathbf{a}_t^{\ast}$
\end{algorithmic}
\end{algorithm}

\noindent (3) \textbf{Objective Design.}
In ActionMoE, we introduce two dedicated loss functions to promote both expert utilization balance and action diversity:

\noindent \textbf{Mixture Balancing Loss} ($\mathcal{L}_{\text{balance}}$):  
This objective encourages uniform expert selection and stabilizes representation learning. It is defined as
\begin{equation}
    \mathcal{L}_{\text{balance}} = \lambda_{\text{aux}} \mathrm{AUX}(\mathbf{p}, \mathbf{u}) + (1 - \lambda_{\text{aux}}) \|\mathbf{h}_t - \mathbf{h}_t^{\text{shared}}\|_2^2
    \label{equ:balance}
\end{equation}
where Auxiliary Loss (AUX)~\cite{wang2024auxiliary} penalizes imbalanced expert usage and $\mathbf{p}$ represents the routing probabilities assigned to each expert, and $\mathbf{u}$ denotes the observed utilization frequency of experts within a mini-batch. The anchoring term $\|\mathbf{h}_t - \mathbf{h}_t^{\text{shared}}\|_2^2$ aligns the hidden state with the shared expert output $\mathbf{h}_t^{\text{shared}} = \operatorname{FFN}_{\text{shared}}(\mathbf{h}_t)$. The set $\lambda_{\text{aux}}$ represents hyperparameters that balance the relative importance of primary and auxiliary loss components.

\noindent \textbf{Action Diversity Loss} ($\mathcal{L}_{\text{div}}$):  
This loss enhances the diversity of exploratory actions by minimizing their similarity to the nominal policy action:
\begin{equation}
    \mathcal{L}_{\text{div}} = \frac{1}{M} \sum_{m=1}^{M} \cos \left( \mathbf{a}_t^{\ast},\, \hat{\mathbf{a}}_t \right)
    \label{equ:div}
\end{equation}
where $\cos(\mathbf{u}, \mathbf{v}) = \frac{\langle \mathbf{u}, \mathbf{v} \rangle}{\|\mathbf{u}\|_2 \|\mathbf{v}\|_2}$ denotes the cosine similarity. Minimizing $\mathcal{L}_{\text{div}}$ encourages the exploratory actions $\mathbf{a}_t^{\ast}$ to be orthogonal to the nominal policy final output $\hat{\mathbf{a}}_t$, thus improving action space coverage and facilitating more effective exploration.

\subsection{Multi-Objective Loss}
\label{sec:loss}
The GRAD framework adopts a multi-objective optimization paradigm, jointly leveraging several complementary learning signals to improve model performance. Specifically, the overall training objective integrates policy learning, value estimation, expert balancing, and action diversity, as defined in Equation~\eqref{eq:loss_policy}, Equation~\eqref{equ:value}, and Equation~\eqref{equ:balance}, Equation~\eqref{equ:div}.Each loss component is designed to target a distinct aspect of the learning process.

The composite loss function is formulated as follows:
\begin{align}
\mathcal{L} = \mathcal{L}_{\text{policy}} + \mathcal{L}_{\text{value}} + \lambda_b\, \mathcal{L}_{\text{balance}} + \lambda_d\, \mathcal{L}_{\text{div}}
\label{eq:total_loss}
\end{align}
where $\lambda_b$ and $\lambda_d$ are hyperparameters that control the relative weighting of the expert balancing and action diversity objectives, respectively.

\section{Experiment}
\subsection{Setup}\label{sec:metrics}

\noindent\textbf{Datasets.} 
Unlike previous work that relies on private bidding logs from closed-source systems, we adopt \textit{AuctionNet}, a publicly available large-scale bidding dataset released by Alibaba~\cite{su2024auctionnet}. To our knowledge, it is the largest public dataset in this domain, containing over 500 million auction records. A sparse variant is also provided to simulate more challenging scenarios. The key statistics are summarized in Table~\ref{tab:statistics} from the Appendix \ref{app:data}.

\noindent\textbf{Evaluation Metrics.} 
Following GAVE~\cite{gao2025generative}, we evaluate performance using a composite score that balances total value and constraint satisfaction. Specifically, we define:
\begin{align}
\text{penalty}_j &= \min\left\{ \left( \frac{C_j}{\sum_i x_i c_{ij} / \sum_i x_i v_{ij}} \right)^{\beta}, 1 \right\} \\
\text{score} &= \left( \sum_i x_i v_i \right) \times \min_{j \in \{1, \ldots, J\}} \text{penalty}_j
\end{align}
Here, $x_i \in \{0,1\}$ indicates whether the $i$-th auction is won, $c_{ij}$ and $v_{ij}$ denote the cost and value (e.g., click or conversion) for constraint $j$, and $C_j$ is the upper bound for the $j$-th KPI constraint (e.g., CPC or CPA). The exponent $\beta = 2$ controls the penalty's sensitivity to constraint violation.

In the AuctionNet dataset, the constraint is CPA, reflecting conversion-based objectives. In practical scenarios, CPC is often used instead. The metric can be easily adapted by substituting $C_j$ and $v_i$ according to the specific KPI constraint.

\subsection{Comparative Evaluation}
We evaluate the performance of \textbf{GRAD} against a range of representative baselines across offline reinforcement learning, diffusion modeling, and auto-bidding strategies:

\begin{itemize}[leftmargin=*]
    \item \textbf{DiffBid}~\cite{guo2024generative}: Models bidding sequences using diffusion processes for trajectory generation.
    \item \textbf{USCB}~\cite{he2021unified}: Online RL-based approach that dynamically adjusts bidding parameters via a unified optimization framework.
    \item \textbf{CQL}~\cite{kumar2020conservative}: Conservative Q-learning algorithm that penalizes overestimation in offline settings.
    \item \textbf{IQL}~\cite{kostrikovoffline}: Utilizes expectile regression to improve policies without relying on value extrapolation.
    \item \textbf{BCQ}~\cite{fujimoto2019off}: Restricts action selection via a learned generative model to remain close to the dataset policy.
    \item \textbf{DT}~\cite{chen2021decision}: Uses a transformer architecture to model decision trajectories via behavior cloning.
    \item \textbf{CDT}~\cite{liu2023constrained}: Offline RL approach for learning safe policies under explicit constraint satisfaction.
    \item \textbf{GAS}~\cite{li2024gas}: Enhances DT with post-training Monte Carlo Tree Search (MCTS) for policy refinement.
    \item \textbf{GAVE}~\cite{gao2025generative}: Introduces constrained exploration and RTG-based budget control within a Causal Transformer framework.
\end{itemize}

\noindent\textbf{Comparison Design.} 
All methods are evaluated on the AuctionNet dataset and its sparse variant under varying budget constraints (50\%, 75\%, 100\%, 125\%, 150\% of the maximum). Performance is measured using the composite score metric described in Section~\ref{sec:metrics}, which balances total value and constraint satisfaction. During training, we utilize the PyTorch on two NVIDIA A100 GPUs. For more detailed hyperparameters, please refer to
Table ~\ref{grad_hyperparameters} from Appendix \ref{app:hyper}. 

\noindent\textbf{Comparison Results.} 
The results in Table~\ref{tab:overall} demonstrate that GRAD consistently achieves superior performance across all budget levels, achieving the highest scores in the majority of experimental configurations. Even in settings where it is marginally outperformed by GAVE, GRAD remains competitive and outperforms all other baseline methods. Methods based on generative modeling, such as DT and CDT, also exhibit strong performance, highlighting their advantage over conventional offline RL approaches like IQL. In contrast, DiffBid underperforms in this large-scale setting, potentially due to the complexity introduced by predicting complete bidding trajectories and modeling inverse dynamics, which may hinder its learning efficiency despite its theoretical benefits.

\begin{table*}[t]
  \centering
  \vspace{-2mm}
  \caption{Performance comparison. The highest score is highlighted in bold, while the second-best result among the baselines is underlined.}
  \vspace{-3mm}
  \setlength\tabcolsep{9pt}
  \scalebox{0.9}{
    \begin{tabular}{ccccccccccccc}
    \toprule
    Dataset & Budget & DiffBid & USCB  & CQL   & IQL   & BCQ   & DT    & CDT   & GAS   & GAVE & GRAD  \\
    \midrule
    \multirow{5}[2]{*}{AuctionNet} & 50\%  & 54    & 86    & 113   & 164   & 190   & 191   & 174   & 193   & \underline{201} & \textbf{204}  \\
          & 75\%  & 100   & 135   & 139   & 232   & 259   & 265   & 242   & 287   & \textbf{296} & \underline{293}   \\
          & 100\% & 152   & 157   & 171   & 281   & 321   & 329   & 326   & 359   & \textbf{376} & \underline{372}    \\
          & 125\% & 193   & 220   & 201   & 355   & 379   & 396   & 378   & 409   & \underline{421} & \textbf{432}  \\
          & 150\% & 234   & 281   & 238   & 401   & 429   & 450   & 433   & 461   & \underline{467} & \textbf{476}  \\
    \midrule
    \multirow{5}[2]{*}{AuctionNet-Sparse} & 50\%  & 9.9   & 11.5  & 12.8  & 16.5  & 17.7  & 14.8  & 11.2  & 18.4  & \underline{19.6} &  \textbf{20.0} \\
          & 75\%  & 15.4  & 14.9  & 16.7  & 22.1  & 24.6  & 22.9  & 18.0  & 27.5  & \underline{28.3} & \textbf{28.5} \\
          & 100\% & 19.5  & 17.5  & 22.2  & 30.0  & 31.1  & 29.6  & 31.2  & 36.1 & \underline{37.2} & \textbf{37.4} \\
          & 125\% & 25.3  & 26.7  & 28.6  & 37.1  & 34.2  & 34.3  & 31.7  & 40.0  & \underline{42.7} &\textbf{43.2}\\
          & 150\% & 30.8  & 31.3  & 35.8  & 43.1  & 37.9  & 44.5  & 39.1  & 46.5  & \underline{47.4} & \textbf{47.5}  \\
    \bottomrule
    \end{tabular}%
    }
  \label{tab:overall}%
  \vspace{-2mm}
\end{table*}%


\subsection{Parameter Analysis}
\begin{figure}[ht]
    \centering
    \vspace{-2mm}
    \includegraphics[width=1\linewidth]{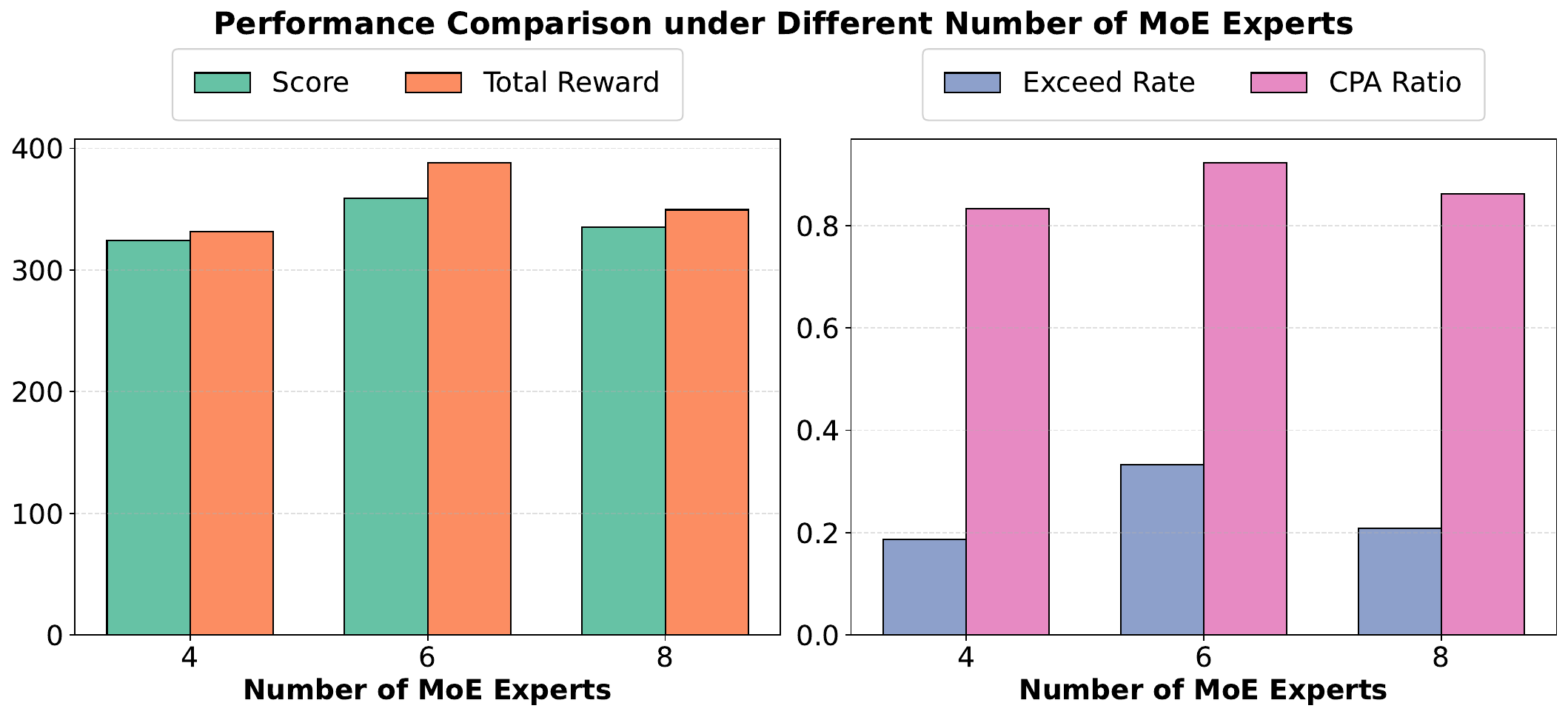}
    \caption{Performance with different numbers of MoE experts. Bar plots show Score and Total Reward (left Y-axis), while the lines indicate Exceed Rate and CPA Ratio (right Y-axis).}
    \label{fig:moe_pa}
\end{figure}

Within the AuctionNet dataset under 100\% budget allocation, we evaluate the impact of Mixture-of-Experts (MoE) configuration on model performance. As shown in Figure~\ref{fig:moe_pa}, four critical metrics—Score, Total Reward, Exceed Rate, and CPA Ratio—are systematically compared across varying expert counts. 

The results show that increasing the number of experts from 4 to 6 improves all metrics, particularly the Exceed Rate and the CPA Ratio. However, performance degrades slightly when increasing to 8 experts. This likely results from increased model complexity and inefficient expert usage causing training instability and gradient dilution, necessitating balanced expert selection for optimal capacity-efficiency trade-offs.




\subsection{Ablation Study}

Table~\ref{tab:ablation} delineates the performance characteristics of GRAD variants under systematic component ablation across varying budget conditions. Comparative analysis 
reveals that the ActionMoE module contributes substantially to performance enhancement, particularly under constrained budget scenarios, by expanding the actionable decision space through its multi-expert architecture. The Value Estimator demonstrates critical importance in high-budget regimes, enabling adaptive constraint-aware optimization that dynamically allocates resources. Under sparse environmental conditions, both components exhibit amplified contributions, with the Value Estimator showing exceptional utility. The integrated GRAD framework achieves optimal performance across all budget levels by leveraging complementary mechanisms: ActionMoE diversifies strategic exploration while the Value Estimator provides contextual constraint guidance, resulting in robust decision-making efficacy.

\begin{table}[t]
  \centering
  \caption{Ablation Study on GRAD Components}
  \vspace{-2mm}
  \setlength{\tabcolsep}{6pt}
  \begin{tabular}{lcccccc}
  \toprule
  \multirow{2}{*}{\textbf{Configuration}} & \multicolumn{3}{c}{\textbf{AuctionNet}} & \multicolumn{3}{c}{\textbf{AuctionNet-Sparse}} \\
  \cmidrule(lr){2-4} \cmidrule(lr){5-7}
  \textbf{Budget} & 50\% & 75\% & 150\% & 50\% & 75\% & 150\% \\
  \midrule
  
  \textbf{GRAD} & \textbf{204} & \textbf{293} & \textbf{476} & \textbf{20.0} & \textbf{28.5} & \textbf{47.5} \\
  \midrule
  
  \textbf{w/o $\mathbf{A}$} 
    & 193 & 274 & 452 & 14.2 & 22.0 & 40.2 \\
    & \footnotesize{(-11)} & \footnotesize{(-19)} & \footnotesize{(-24)} 
    & \footnotesize{(-5.8)} & \footnotesize{(-6.5)} & \footnotesize{(-7.3)} \\
  \addlinespace[0.2em]
  
  \textbf{w/o $\mathbf{V}$} 
    & 195 & 278 & 466 & 16.3 & 20.1 & 42.2 \\
    & \footnotesize{(-9)} & \footnotesize{(-15)} & \footnotesize{(-10)} 
    & \footnotesize{(-3.7)} & \footnotesize{(-8.4)} & \footnotesize{(-5.3)} \\
  \addlinespace[0.2em]
  
  \textbf{w/o ($\mathbf{A}$ \& $\mathbf{V}$)} 
    & 191 & 265 & 433 & 11.2 & 18.0 & 39.1 \\
    & \footnotesize{(-13)} & \footnotesize{(-28)} & \footnotesize{(-43)} 
    & \footnotesize{(-8.8)} & \footnotesize{(-10.5)} & \footnotesize{(-8.4)} \\
  \midrule
  \end{tabular}
  \vspace{1mm}
\begin{minipage}{\textwidth}
  \footnotesize
  \textit{Note:} 
  $\mathbf{A}$: ActionMoE module; $\mathbf{V}$: Value Estimator module \\
  \hspace*{2.4em}\textbf{w/o}: Ablation configuration (component removal)
\end{minipage}
  \vspace{-8mm}

  \label{tab:ablation}
\end{table}

\section{Online A/B Test}
\label{sec:abtest}
\subsection{Setup}
To evaluate the online performance of the proposed GRAD, we collected bidding logs from Meituan's advertising platform between January and March 2025. The dataset comprises two scales: a base set of 5,000 trajectories and an extended set of 30,000 trajectories. Each trajectory aggregates sequential bid decisions using PID control logic and contains timestamped bid requests sampled from real production traffic. The dataset captures temporal bidding patterns at 15-minute intervals over daily cycles and bid adjustments ranging from 80\% to 120\% of base prices, providing realistic variability for offline RL training while adhering to practical business constraints. The key elements are summarized as follows:
\begin{itemize}[leftmargin=*]
    \item \textbf{State}: Campaign features including bdudget, cost, charge, price, time-based budget allocation, time-based cost speed, predicted conversion rates, real CPC, etc.
    \item \textbf{Action}: Bid value $a_t$ at each timestep.
    \item \textbf{Reward}: Composite reward balancing user engagement and constraint enforcement:
    \begin{equation}
    \label{eq:reward}
    Reward = \log\big(1 + 1000 \times \mathrm{CTR}\big) - \lambda \cdot \min\left(P_{\max}, \left( \frac{\mathrm{CPC} - \vartheta}{\vartheta} \right)^3 \right)
    \end{equation}
    where $\lambda \in \{0,1\}$ activates the penalty when constraints are violated, $\vartheta$  the CPC threshold, and $P_{\max}$ limits the penalty magnitude.
\end{itemize}

\subsection{Deployment}

\begin{figure}[ht]
    \centering
    \vspace{-2mm}
    \includegraphics[width=0.98\linewidth]{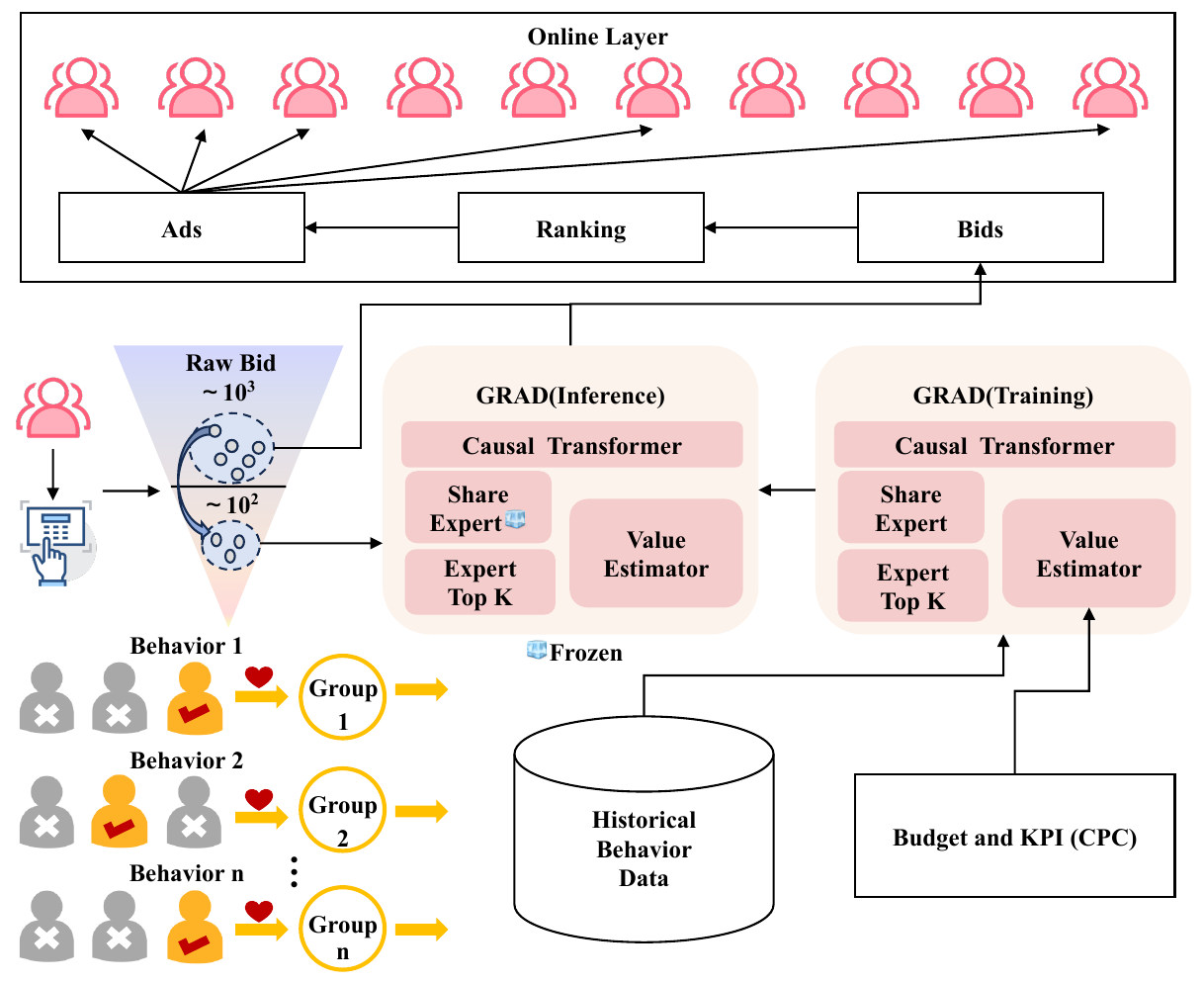}
    \caption{Overview of the Online Auto-bidding System}
    \label{fig:deployment}
     \vspace{-2mm}
\end{figure}

We deploy the GRAD in Meituan’s advertising system under the Multiple Constraint Bidding (MCB) scenario. Under the MCB setting, advertisers specify budgets along with optional CPC or ROI constraints, and the bidding strategy aims to maximize conversions while satisfying these constraints. As illustrated in Figure~\ref{fig:deployment}, the deployment pipeline consists of two main phases:
\begin{itemize}[leftmargin=*]
\item \textbf{Training.} Historical users' behavior data are aggregated into groups and serve as inputs. The model optimizes bidding parameters under budget and KPI constraints (e.g., CPC). The architecture is based on a Causal Transformer with two key components: (1) ActionMoE, which integrates a shared expert for general feature extraction and domain-specific routed experts with top-\(K=1\) routing, and (2) Value Estimator, which enforces joint budget-KPI constraint adherence.
\item \textbf{Inference.} To ensure online efficiency, the shared expert is frozen, and only the most relevant routed expert (\(K=1\)) is activated per request, allowing low-latency bidding. Incoming user requests undergo initial deduplication, filtering roughly \(\sim\!10^3\) raw bid candidates down to \(\sim\!10^2\) qualified ads. These candidates are then processed by the model using compressed ad features and user context, producing optimized bids.
\end{itemize}

Finally, the generated bids are passed to the ranking module for the final ad selection.


\vspace{-2mm}
\subsection{Online A/B Test Results}
The online A/B test demonstrates GRAD's significant performance improvements across key advertising metrics. Our online A/B testing is conducted for 7 full days, as in Table~\ref{tab:online}. 
Complementary gains in CTR (+3.93\%), CPC\_CR(5.64\%), ROI (+10.68\%), and GMV (+2.18\%) 
confirm GRAD's balanced optimization capabilities  against the best RL solution in Meituan's advertising platform. The CPC\_CR metric quantifies the temporal reliability of cost-efficiency control, formally defined as:
\vspace{-1mm}
\begin{equation}
\label{eq:cpc_cr}
\text{CPC\_CR} = \frac{1}{T} \sum_{t=1}^{T} \mathbb{I}\left( \overline{\text{CPC}}_t \leq \gamma C_{\text{target}} \right) \times 100\% 
\end{equation}
where $\overline{\text{CPC}}_t$ is the daily average CPC on day $t$ with $\mathcal{I}_t$ being impressions acquired on day $t$, $\mathbb{I}(\cdot)$ is the indicator function, $C_{\text{target}}$ is the contractual CPC target, and $\gamma=1.2$ is the tolerance factor permitting 20\% operational flexibility. This measure evaluates the system's ability to maintain cost discipline under auction environments.

\vspace{-2mm}
\begin{table}[h!]
\caption{Live experiment: Online A/B test results}
\vspace{-3mm}
\begin{tabular}{lcccc}
\toprule
\textbf{Method} & \textbf{CTR} & \textbf{CPC\_CR} & \textbf{ROI} & \textbf{GMV} \\ 
\midrule
GRAD  & +3.93\% & +5.64\% & +10.68\% & +2.18\% \\
\bottomrule
\end{tabular}
\vspace{-2mm} 

\label{tab:online}
\vspace{-4mm}
\end{table}

\section{Related Work}
\subsection{Offline Reinforcement Learning}


Offline reinforcement learning (RL) learns optimal policies from static datasets~\cite{schweighofer2022dataset,lepak2024reinforcement}, eliminating costly/dangerous online interaction. Unlike traditional RL requiring active exploration, it extracts behaviors solely from pre-collected data~\cite{ball2023efficient,levine2020offline}, proving invaluable in exploration-constrained domains.

Recent advances in this field have introduced innovative techniques to address key challenges such as distributional shift and overestimation bias~\cite{aggarwal2025multi}. Conservative Q-learning (CQL)~\cite{kumar2020conservative} mitigates value overestimation by penalizing out-of-distribution actions, while Batch-Constrained Q-learning (BCQ)~\cite{fujimoto2019off} restricts policy updates to actions supported by the dataset. Implicit Q-learning (IQL) ~\cite{kostrikovoffline} further improves stability by learning value functions without explicit policy constraints. Additionally, Decision Transformer~\cite{chen2021decision} leverages sequence modeling to capture long-range dependencies in offline data, and IDQL~\cite{hansen2023idql} integrates diffusion models with implicit Q-learning to enhance policy robustness. These approaches collectively advance the scalability and reliability of offline RL in complex, real-world settings.

\subsection{Auto-bidding}
Auto-bidding systems have become a cornerstone of programmatic advertising~\cite{he2024hierarchical}, enabling automated real-time bidding~\cite{sayedi2018real,liu2022real,cai2025rtbagent} (RTB) for ad placements while optimizing key performance indicators (KPIs) such as click-through rates (CTR), conversions, or return on ad spend (ROAS) under strict budget constraints. These systems leverage machine learning and optimization techniques  to adjust bids in complex, competitive auction environments dynamically.

Reinforcement learning (RL) has gained significant traction in optimizing auto-bidding strategies~\cite{korenkevych2024offline, li2024trajectory, chen2023model}. The field has evolved from early simulated bidding environments to~\cite{he2021unified}'s unified approach handling multiple constraints, followed by~\cite{jin2018real}'s multi-agent competitive settings and~\cite{mou2022sustainable}'s performance improvements over rule-based strategies. Most recently,~\cite{wen2022cooperative} deployed a cooperative multi-agent framework incorporating richer market information. Despite demonstrated successes, these RL approaches continue to face challenges,  including sample inefficiency, training instability, and adaptation difficulties in dynamic auction environments.

\subsection{Generative Models}
Generative models aim to capture the underlying data distribution of a given dataset or model conditional probability distributions between variables. Recently, transformer-based architectures have been adapted to model complex DMP through autoregressive generative frameworks, as demonstrated in models like GAVE~\cite{gao2025generative} and GAS~\cite{li2024gas}. Concurrently, diffusion models have emerged as a powerful paradigm for conditional sample generation. These models iteratively refine samples via a reverse denoising process conditioned on inputs, with representative examples including DiffBid~\cite{guo2024generative}.


\section{Conclusion}

In this paper, we address the growing complexity and diversity of advertiser demands in the online advertising market by proposing \textsc{GRAD}, a novel generative foundation model for automated bidding. Our approach leverages a Mixture-of-Experts (MoE) module for efficient exploration of the action space and a Causal Transformer-based value estimator for precise reward evaluation under a variety of constraints. Experimental results on large-scale real-world datasets demonstrate that GRAD not only achieves superior adaptability to diverse advertiser objectives and constraints but also delivers robust performance in dynamic and challenging environments. This work highlights the potential of scalable generative models in advancing the intelligence and flexibility of auto-bidding systems, paving the way for future research on universal, highly adaptive, and industrially deployable advertising solutions. Notably, GRAD has already been deployed in the production environment of Meituan, a large-scale commercial platform, demonstrating its robustness and adaptability in real-world applications.

\bibliographystyle{ACM-Reference-Format}
\bibliography{sample-base}
\appendix

\section{Datset Details}
\label{app:data}
The Dataset statistics are summarized in Table~\ref{tab:statistics}.
\begin{table}[H]
  \centering
  \caption{Data statistics}
  \vspace{-4mm}
  \scalebox{0.9}{
    \begin{tabular}{lll}
    \toprule
    Params & AuctionNet & AuctionNet-Sparse \\
    \midrule
    Trajectories & 479,376 & 479,376 \\
    Delivery Periods & 9,987 & 9,987 \\
    Time steps in a trajectory & 48    & 48 \\
    State dimension & 16    & 16 \\
    Action dimension & 1     & 1 \\
    Return-To-Go Dimension & 1     & 1 \\
    Action range & [0, 493] & [0, 589] \\
    Impression's value range & [0, 1] & [0, 1] \\
    CPA range & [6, 12] & [60, 130] \\
    Total conversion range & [0, 1512] & [0, 57] \\
    \bottomrule
    \end{tabular}%
    }
  \label{tab:statistics}%
  \vspace{-2mm}
\end{table}%

\section{Hyperparameters Setting}
\label{app:hyper}
The hyperparameter details are summarized in Table \ref{grad_hyperparameters}.

\begin{table}[H]
\caption{The detailed hyperparameters of GRAD}
\label{grad_hyperparameters}
\scalebox{0.9}{
\begin{tabular}{ll}
\toprule
Hyperparameters     & Value  \\ 
\midrule
Batch size          & 128    \\
Number of steps     & 400000 \\
Sequence length     & 20     \\
Learning rate       & 1e-5   \\
Number of attention layers       & 8   \\
Number of heads       & 16   \\
Optimizer           & AdamW  \\
Optimizer eps       & 1e-8   \\
Weight decay        & 1e-2   \\
Scale               & 2000   \\
Episode length      & 48     \\
Hidden size         & 512    \\
Activation function & ReLU   \\
Gamma               & 0.99   \\
Tau                 & 0.01   \\
Expectile           & 0.7    \\ 
Number of Experts   & 6      \\ 
$\lambda_{\text{aux}} $  & 0.2      \\ 
\bottomrule
\end{tabular}}
\end{table}
\section{Theory of Action-Space Design under Hard Constraints}
\subsection{Preliminaries: CMDP and Feasible Action Sets}
We consider a constrained Markov decision process (CMDP) with tuple $(\mathcal{S},\mathcal{A},P,r,\gamma)$ and a set of $K$ hard constraints represented by measurable functions $C_k:\mathcal{S}\times \mathcal{A}\to \mathbb{R}$, $k\in\{1,\dots,K\}$. An action $a\in\mathcal{A}$ is \emph{feasible} at state $s$ if $C_k(s,a)\le 0$ for all $k$. Define the state-dependent feasible set
\[
\mathcal{F}(s)\;=\;\{a\in\mathcal{A}:\; C_k(s,a)\le 0,\;\forall k\}.
\]
For auto-bidding, typical hard constraints include instantaneous or per-interval budget pacing and delivery limits; soft constraints like CPC/CPA can be embedded via penalties in the objective.
We denote by $a_{t-1}\in \mathcal{F}(s_t)$ a previously executed feasible action at time $t-1$ evaluated at state $s_t$, and consider trust-region style perturbations of $a_{t-1}$ to construct candidate actions.
\begin{assumption}[Constraint Lipschitzness in Action]
\label{ass:lipschitz}
For each $k$, $C_k(s,\cdot)$ is $L_k$-Lipschitz on $\mathcal{A}$ for all $s$, i.e.,
\[
|C_k(s,a)-C_k(s,a')|\;\le\; L_k\,\|a-a'\|\quad\forall a,a'\in\mathcal{A},\;\forall s\in\mathcal{S}.
\]
Let $L:=\max_k L_k$.
\end{assumption}
\begin{assumption}[Feasibility Margin]
\label{ass:margin}
There exists a \emph{margin} $\eta>0$ such that for the current state $s_t$ and baseline action $a_{t-1}$,
\[
\max_{k} C_k(s_t,a_{t-1})\;\le\; -\eta.
\]
\end{assumption}
\subsection{Feasibility Preservation under Trust-Region Perturbations and Residuals}
ActionMoE generates candidate actions by element-wise scaling $a_{t-1}$: $a' = a_{t-1}\odot f$, with $f\in [1-\varepsilon,1+\varepsilon]^d$, and then adds a bounded residual $U_t$ from a Lipschitz MLP to form $a^\star = a'+U_t$.
\begin{lemma}[Norm Bound for Elementwise Scaling]
\label{lem:scale}
Let $f\in[1-\varepsilon,1+\varepsilon]^d$ and $a'\!=\!a\odot f$. Then
\[
\|a'-a\|\;\le\;\varepsilon\,\|a\|.
\]
\end{lemma}
\begin{proof}
By triangle inequality and elementwise bounds,
\(
\|a'-a\|^2=\sum_{i=1}^d (f_i-1)^2 a_i^2\le \varepsilon^2 \sum_{i=1}^d a_i^2=\varepsilon^2\|a\|^2.
\)
\end{proof}
\begin{proposition}[Feasibility Preservation under Trust-Region Scaling]
\label{prop:feasible-scaling}
Under Assumptions \ref{ass:lipschitz}--\ref{ass:margin}, let $a' = a_{t-1}\odot f$ with $f\in[1-\varepsilon,1+\varepsilon]^d$ and $\varepsilon\le \eta/(L\|a_{t-1}\|)$. Then $a'\in \mathcal{F}(s_t)$, i.e., $C_k(s_t,a')\le 0$ for all $k$.
\end{proposition}
\begin{proof}
By Lipschitzness and Lemma \ref{lem:scale},
\[
C_k(s_t,a') \le C_k(s_t,a_{t-1}) + L_k \|a'-a_{t-1}\|
\le -\eta + L\,\varepsilon \|a_{t-1}\|\le 0.
\]
\end{proof}
\begin{assumption}[Residual Bound and MLP Lipschitzness]
\label{ass:residual}
The residual $U_t=\mathrm{MLP}(Y_t)$ satisfies $\|U_t\|\le \delta$, with $\mathrm{MLP}$ $\nu$-Lipschitz and inputs $Y_t$ bounded. Moreover, $\delta$ is chosen such that $\delta \le \eta/L - \varepsilon\|a_{t-1}\|$.
\end{assumption}
\begin{proposition}[Feasibility Preservation with Residual Fusion]
\label{prop:feasible-residual}
Under Assumptions \ref{ass:lipschitz}--\ref{ass:residual}, let $a^\star = a' + U_t$ where $a'$ as in Prop.~\ref{prop:feasible-scaling}. Then $a^\star\in \mathcal{F}(s_t)$.
\end{proposition}

\begin{proof}
By Lipschitzness,
\[
C_k(s_t,a^\star) \le C_k(s_t,a') + L_k \|U_t\| \le 0 + L\,\delta \le 0.
\]
\end{proof}

\begin{corollary}[Projection Guarantees]
\label{cor:projection}
If residual bounds cannot be guaranteed, define the projection operator $\Pi_{\mathcal{F}(s_t)}(a):=\arg\min_{b\in\mathcal{F}(s_t)}\|b-a\|$. Then for any candidate $\tilde a$, the projected action $a^\dagger=\Pi_{\mathcal{F}(s_t)}(\tilde a)$ is feasible: $a^\dagger\in\mathcal{F}(s_t)$. Moreover, if $C_k$ are convex in $a$, $\Pi_{\mathcal{F}(s_t)}$ is non-expansive:
\(
\|\Pi_{\mathcal{F}}(x)-\Pi_{\mathcal{F}}(y)\|\le \|x-y\|.
\)
\end{corollary}

\subsection{Monotone Improvement under Penalized Objective}
Define a \emph{penalized} objective for policy $\pi$:
\[
J_\lambda(\pi):=\mathbb{E}_\pi\Big[\sum_{t=0}^\infty \gamma^t\big(r(s_t,a_t)-\lambda\,\phi(C(s_t,a_t))\big)\Big],
\]
where $C(s,a):=\max_k C_k(s,a)$, $\phi:\mathbb{R}\to\mathbb{R}_+$ is convex, non-decreasing with $\phi(x)=0$ for $x\le 0$, and $\lambda>0$.
Let $Q_\lambda^\pi(s,a)$ be the penalized action-value function and $A_\lambda^\pi(s,a):=Q_\lambda^\pi(s,a)-V_\lambda^\pi(s)$ the penalized advantage.
\begin{lemma}[Performance Difference for Penalized CMDP]
\label{lem:perf-diff}
For any policies $\pi,\pi'$, under standard regularity,
\[
J_\lambda(\pi')-J_\lambda(\pi)
=\frac{1}{1-\gamma}\,\mathbb{E}_{s\sim d_{\pi'}}\Big[\mathbb{E}_{a\sim \pi'(\cdot|s)}A_\lambda^\pi(s,a)\Big],
\]
where $d_{\pi'}$ is the discounted occupancy measure under $\pi'$.
\end{lemma}
\begin{proof}
Apply the performance difference lemma to the modified reward $r_\lambda:=r-\lambda\,\phi\circ C$.
\end{proof}
\begin{assumption}[Trust-Region Policy Update]
\label{ass:trust-region}
Consider an update $\pi\to\pi'$ satisfying for all $s$:
\[
\mathrm{KL}(\pi'(\cdot|s)\|\pi(\cdot|s))\le \epsilon_{\mathrm{KL}},
\quad
\mathbb{E}_{a\sim \pi'(\cdot|s)}A_\lambda^\pi(s,a)\ge \alpha\,\mathbb{E}_{a\sim \pi(\cdot|s)}A_\lambda^\pi(s,a)
\]
for some $\alpha\in(0,1]$.

\end{assumption}
\begin{proposition}[Monotone Improvement Lower Bound]
\label{prop:monotone}
Under Lemma \ref{lem:perf-diff} and Assumption \ref{ass:trust-region}, suppose $|A_\lambda^\pi(s,a)|\le B$ and the density ratio $\frac{d_{\pi'}(s)}{d_{\pi}(s)}\ge \rho_{\min}>0$ for states with non-zero occupancy. Then
\[
J_\lambda(\pi')-J_\lambda(\pi)\;\ge\;\frac{\rho_{\min}}{1-\gamma}\Big(\alpha\,\mathbb{E}_{s\sim d_{\pi}}\mathbb{E}_{a\sim \pi(\cdot|s)}A_\lambda^\pi(s,a)\Big)\;-\;\mathcal{O}(\epsilon_{\mathrm{KL}}\,B).
\]
In particular, for sufficiently small $\epsilon_{\mathrm{KL}}$, the update yields a positive improvement whenever the baseline expected penalized advantage is positive.
\end{proposition}
\begin{proof}
Combine the performance difference identity with occupancy ratio control and Pinsker's inequality to bound deviations induced by the trust-region constraint.
\end{proof}

\subsection{Action Diversity and Coverage Guarantees}
ActionMoE employs an action diversity loss that encourages low cosine similarity between the final nominal action $\hat a_t$ and exploratory actions $a^\star_{t,m}$:
\[
\frac{1}{M}\sum_{m=1}^M \cos(\hat a_t,a^\star_{t,m}) \le \rho,
\qquad \rho\in[0,1).
\]
Equivalently, pairwise angular separation $\theta_m:=\arccos\big(\cos(\hat a_t,a^\star_{t,m})\big)\ge \arccos(\rho)$.
\begin{proposition}[Gram Matrix Well-Conditioning via Angular Separation]
\label{prop:gram}
Let $v_0:=\hat a_t/\|\hat a_t\|$ and $v_m:=a^\star_{t,m}/\|a^\star_{t,m}\|$ in $\mathbb{R}^d$, $m=1,\dots,M$, with $\langle v_0,v_m\rangle \le \rho$. Form the matrix $V=[v_1,\dots,v_M]\in\mathbb{R}^{d\times M}$ and its Gram matrix $G:=V^\top V\in\mathbb{R}^{M\times M}$. Then for any unit vector $x\in\mathbb{R}^M$,
\[
x^\top G x = \Big\|\sum_{m=1}^M x_m v_m\Big\|^2 \;\ge\; (1-\rho^2)\,\|x\|^2\;-\;\sum_{m\neq n}|x_m x_n|\,\Delta_{mn},
\]
where $\Delta_{mn}:=|\langle v_m - \rho v_0, v_n - \rho v_0\rangle|$. In particular, if the $v_m$ are additionally mutually separated such that $|\langle v_m,v_n\rangle|\le \kappa<1$ for $m\neq n$, then
\[
\lambda_{\min}(G)\;\ge\;1-\kappa(M-1),
\]
and the exploration set is well-conditioned provided $\kappa<\frac{1}{M-1}$.

\end{proposition}

\begin{proof}
Decompose each $v_m$ into parallel/orthogonal components to $v_0$ and apply Gershgorin bounds for the Gram matrix under mutual separation.
\end{proof}
\begin{corollary}[Coverage Radius on Unit Sphere]
\label{cor:coverage}
If each $v_m$ satisfies $\langle v_0,v_m\rangle\le \rho$, then the spherical caps $\mathcal{C}(\theta)$ of half-angle $\theta=\arccos(\rho)$ centered at $v_m$ lie outside the cap centered at $v_0$. Hence the exploratory set covers directions outside a cap of measure proportional to $(1-\rho)$; decreasing $\rho$ expands directional coverage.
\end{corollary}
\subsection{Sparse MoE Routing and Gradient Stability}
ActionMoE uses a shared expert $F_{\mathrm{shared}}$ and a single routed expert $F_{m^\star}$ selected by top-1 gating, with output
\[
Y_t = \mathrm{LN}\big(F_{\mathrm{shared}}(h_t)+F_{m^\star}(h_t)\big).
\]
Assume both FFNs are $L_s,L_r$-Lipschitz and layer norm is $L_{\mathrm{LN}}$-Lipschitz in the region of interest.
\begin{assumption}[Bounded Hidden Norms]
\label{ass:hidden}
There exists $H>0$ such that $\|h_t\|\le H$ almost surely during training, and gradients $\nabla_\theta h_t$ are bounded by $G_h$.
\end{assumption}
\begin{lemma}[Output Lipschitzness]
\label{lem:lips-output}
Under Assumption \ref{ass:hidden}, the composite mapping $h_t\mapsto Y_t$ is $L_Y$-Lipschitz with
\[
L_Y \;\le\; L_{\mathrm{LN}}\,(L_s+L_r).
\]
\end{lemma}
\begin{proof}
By triangle inequality and Lipschitz composition.
\end{proof}
\begin{proposition}[Gradient Norm Bound under Top-1 Routing]
\label{prop:grad-stability}
Let $\theta$ collect parameters of both experts and LN. Under Assumptions \ref{ass:hidden} and Lemma \ref{lem:lips-output}, the gradient of the loss $\mathcal{L}$ w.r.t.\ $\theta$ satisfies
\[
\|\nabla_\theta \mathcal{L}\|\;\le\; C\,\big(\|\nabla_{Y_t}\mathcal{L}\|\cdot L_Y\cdot G_h\big),
\]
for some constant $C$ depending on the architecture depth. In particular, activating only one routed expert (top-1) ensures the Lipschitz constant scales with $(L_s+L_r)$ rather than the sum over all experts, reducing gradient variance and improving stability compared to multi-expert simultaneous activation.
\end{proposition}
\begin{proof}
Chain rule over the computation graph: with only one routed branch active, the pathwise product of Lipschitz constants excludes inactive experts, yielding the stated bound.
\end{proof}

\section{In-depth Analysis}
\begin{figure}[ht]
\centering
\subfloat[Gradient-norm over training steps. The penalized objective yields smoother, lower-variance gradients.]{%
\includegraphics[width=0.48\columnwidth]{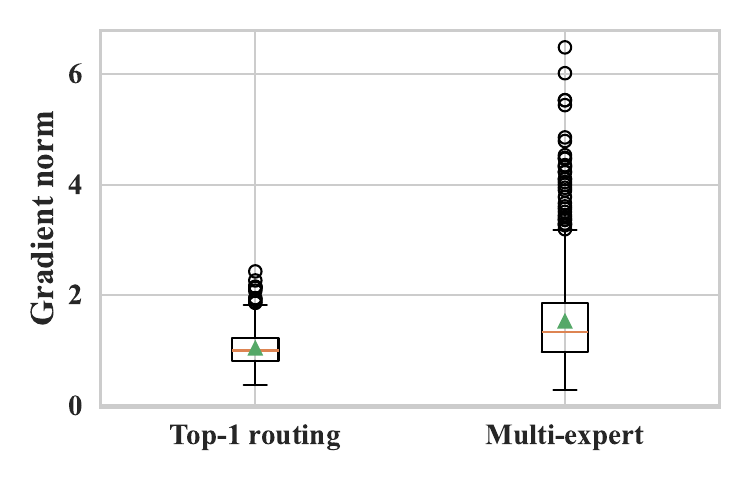}%
\label{fig:grad-norm-compare}}
\hfill
\subfloat[penalized objective $J_{\lambda}$ versus training steps, showing steady improvement and earlier convergence.]{%
\includegraphics[width=0.48\columnwidth]{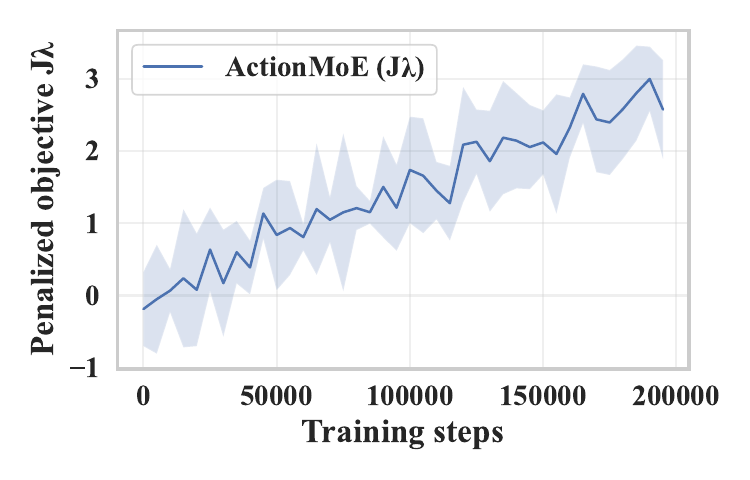}%
\label{fig:penalized-return-curve}}
\caption{Training with the penalized objective ($J_{\lambda}$) improves stability and accelerates convergence.}
\label{fig:grad-penalized-side-by-side}
\end{figure}

Here we compare training dynamics with and without the penalized objective. As shown in Fig.~\ref{fig:grad-penalized-side-by-side}\subref{fig:grad-norm-compare}, the gradient-norm trajectory exhibits smoother updates with fewer large spikes when adding the penalty term ($J_{\lambda}$), indicative of a more stable optimization process. Fig.~\ref{fig:grad-penalized-side-by-side}\subref{fig:penalized-return-curve} plots the penalized objective $J_{\lambda}$ over training steps; we observe a steady improvement and earlier stabilization when using the penalized objective, consistent with reduced constraint violations.

\begin{figure}[ht]
    \centering
    \vspace{-5mm}
    \includegraphics[width=0.98\linewidth]{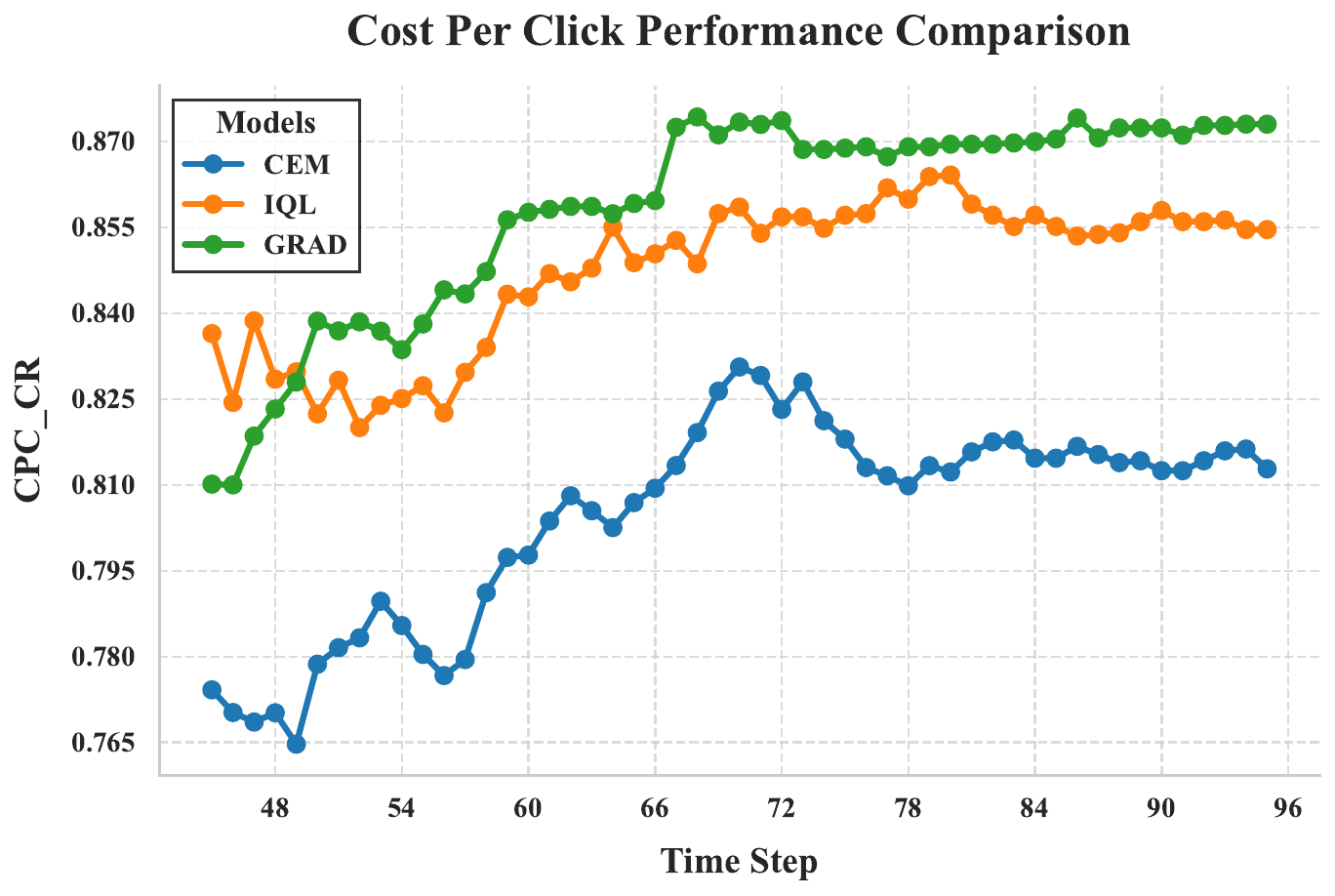}
    \caption{Comparison of CPC\_CR on a randomly trajectory }
    \label{fig:case}
     \vspace{-3mm}
\end{figure}

Figure~\ref{fig:case} presents a randomly sampled trajectory segment of CPC\_CR (CPC achievement rate) over time for three bidding policies: GRAD, IQL, and CEM. Here, CEM denotes a Cross-Entropy Method–based bidding algorithm used for stochastic action optimization. GRAD achieves the highest performance and the greatest stability (lowest variance), outperforming IQL and CEM.


\end{document}